\documentclass{article}

     \usepackage[preprint]{neurips_2021}

\usepackage[utf8]{inputenc} 
\usepackage[T1]{fontenc}    
\usepackage{hyperref}       
\usepackage{url}            
\usepackage{booktabs}       
\usepackage{amsfonts}       
\usepackage{nicefrac}       
\usepackage{microtype}      
\usepackage{xcolor}         

\usepackage{amsmath}
\usepackage{amssymb}
\usepackage{mathtools}
\usepackage{amsthm}
\usepackage{bbm}

\usepackage{microtype}
\usepackage{graphicx}
\usepackage{subfigure}
\usepackage{booktabs} 
\usepackage{graphicx}
\graphicspath{ {./figures/} }
\usepackage{algorithm}

\usepackage{slashbox}  
\usepackage{multirow}
\usepackage{newtxmath}

\usepackage{array}
\usepackage{dsfont}
\usepackage{wrapfig}
\usepackage{times}
\usepackage{soul}
\usepackage{makecell}

\usepackage[capitalize,noabbrev]{cleveref}

\theoremstyle{plain}

\theoremstyle{definition}

\theoremstyle{remark}

\usepackage{pgfplots}
\usetikzlibrary{plotmarks}
\usepgfplotslibrary{fillbetween}
\pgfplotsset{compat=1.15}
\pgfplotsset{
    discard if not/.style 2 args={
        x filter/.code={
            \edef\tempa{\thisrow{#1}}
            \edef\tempb{#2}
            \ifx\tempa\tempb
            \else
                
            \fi
        }
    }
}

\usepackage{algorithmic}
\usepackage{xspace}
\usepackage{color}
\newcommand*{\modelname}{\textsf{SwARo}\@\xspace}
\newcommand*{\eg}{\emph{e.g.},\@\xspace}
\newcommand*{\ie}{\emph{i.e.},\@\xspace}

\DeclareMathOperator*{\argmin}{arg\,min}

\newcommand\Tau{\mathcal{T}}

\title{Adversarial Contrastive Learning by Permuting Cluster Assignments}

\author{%
  Muntasir Wahed \\
  Department of Computer Science\\
  Virginia Tech\\
  \texttt{mwahed@vt.edu}\\
  \And
  Afrina Tabassum \\
  Department of Computer Science\\
  Virginia Tech\\
  \texttt{afrina@vt.edu}\\
   \And
  	Ismini Lourentzou  \\
  Department of Computer Science\\
  Virginia Tech\\
  \texttt{ilourentzou@vt.edu}\\
}

\begin{document}

\maketitle

\begin{abstract}
Contrastive learning has gained popularity as an effective self-supervised representation learning technique. Several research directions improve traditional contrastive approaches, \eg prototypical contrastive methods better capture the semantic similarity among instances and reduce the computational burden by considering cluster prototypes or cluster assignments, while adversarial instance-wise contrastive methods improve robustness against a variety of attacks. To the best of our knowledge, no prior work jointly considers robustness, cluster-wise semantic similarity and computational efficiency.
In this work, we propose \modelname, an adversarial contrastive framework that incorporates cluster assignment permutations to generate representative adversarial samples. We evaluate \modelname on multiple benchmark datasets and against various white-box and black-box attacks, obtaining consistent improvements over state-of-the-art baselines.
\end{abstract}

\section{Introduction}\label{sec:introduction}
Contrastive learning methods aim to learn representations without relying on semantic annotations of instances, by bringing closer augmented samples from the same instance (considered as a positive pair) and pushing further apart samples from other instances (treated as negative samples). Contrastive methods have
gained popularity in recent literature as pretraining methods to improve label efficiency and model performance~\cite{oord2018representation,hjelm2018learning, he2020momentum,chen2020simclr,Caron2020Cluster, chuang2020debiased,henaff2020data,tian2019contrastive, robinson2020hard}.

A central step is the selection of positive and negative examples when computing the contrastive loss. In particular, prior works have shown that large batches of negative samples improve the learned representations~\cite{he2020momentum,chen2020improved}. Computing all pairwise comparisons or utilizing large memory banks of randomly selected negative examples are, however, impractical in realistic applications.
Prior work also shows that instance-wise contrastive learning methods do not account for any global semantic similarities observed in data~\cite{Li2020PCL}.
To address these limitations, a few methods sample representative negative examples via clustering techniques, \eg by contrasting prototypes of varying granularity~\cite{Li2020PCL}, predicting cluster labels with a swapped prediction mechanism~\cite{Caron2020Cluster}, or sampling positive
and negative pairs from weak cluster assignments \cite{sharma2020clustering}. In general, clustering has been an effective way to capture semantic patterns in the data and has been used in several works, \eg in semi-supervised learning~\cite{zhang2009prototype,kuo2020featmatch,li2021cross}, cross-modal learning ~\cite{hu2021learning,alwassel2020self}, out-of-distribution detection and novel class discovery~\cite{yang2021semantically,han2019learning}.

\begin{wrapfigure}{r}{0.5\textwidth}
    \centering
    \includegraphics[width=0.3\textwidth]{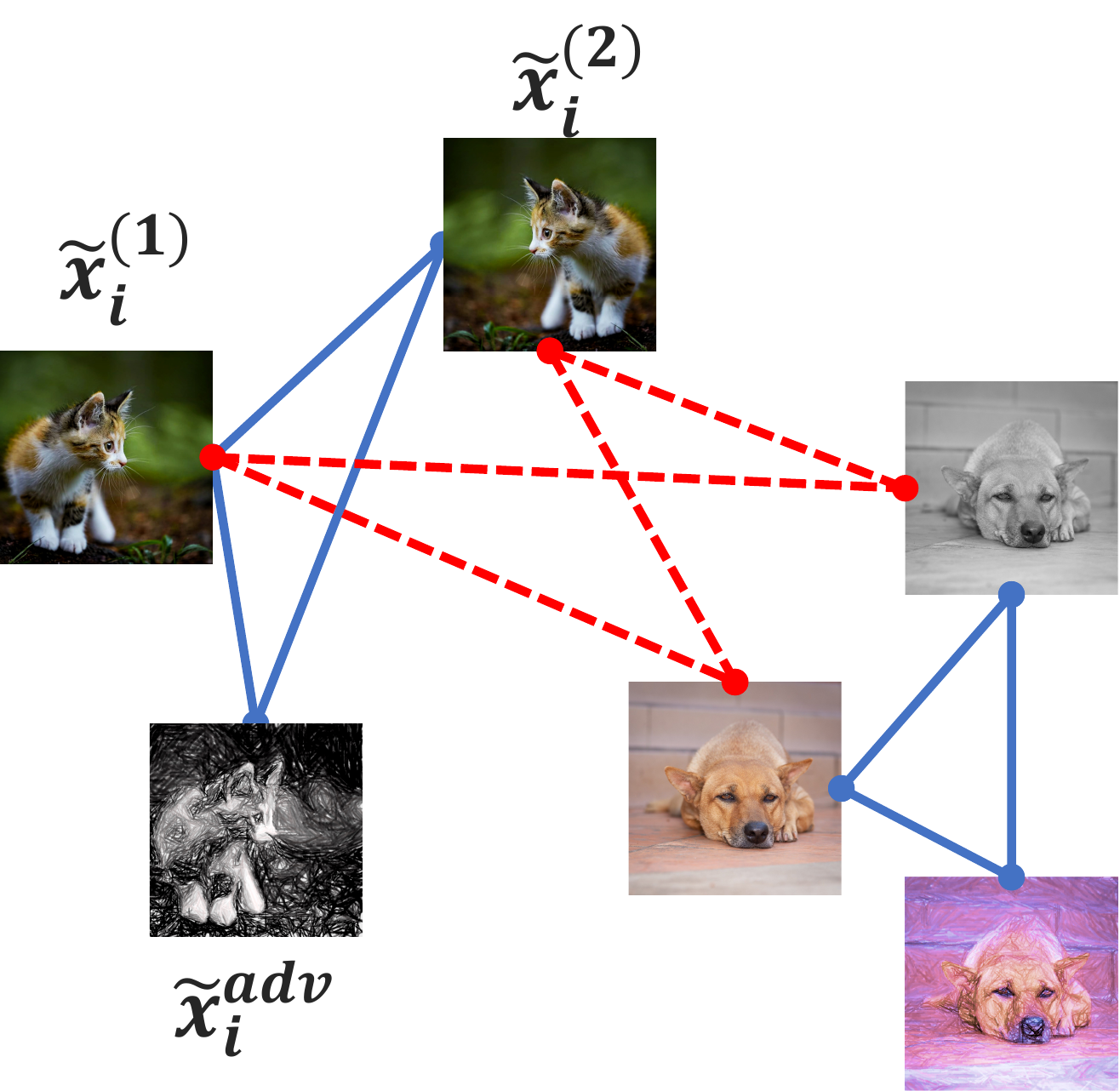}
    \caption{{Prior works maximize the self-supervised contrastive loss between positive examples $\tilde{x}_i^{(1)}, \tilde{x}_i^{(2)}$ when generating adversarial examples $\tilde{x}_i^{adv}$ (\textcolor{blue}{\textbf{blue} } lines). In contrast, \modelname maximizes the loss for positive pairs (\textcolor{blue}{\textbf{blue}} lines) but also creates targeted attacks by minimizing the loss for negative pairs (\textcolor{red}{\textbf{red}} lines).
    }}
    \label{fig:sketch}
\end{wrapfigure}

Nevertheless, most instance-wise and cluster-based methods presented lack robustness to adversarial examples. A few works target improving robustness in contrastive learning by devising adversarial perturbations of positive and negative examples~\cite{Kim2020AdversarialCL,Jiang2020PretrainingACL,mcdermott2021adversarial,bui2021understanding}. These efforts try to design adversarially robust self-supervised learning frameworks by generating perturbations that maximize the contrastive loss of augmented instance-level samples, causing the model to make incorrect predictions. However, semantic cluster structures encoded in the data are rarely considered, and the designed perturbations may not generalize well in downstream tasks. 
Intuitively, and in the absence of label (class) information, the self-supervised gradient-based adversarial attacks imposed can be considered untargeted, as the gradient points towards a direction that maximizes the contrastive loss for same-class instances, without any target class in mind (see Figure~\ref{fig:sketch} \textcolor{blue}{\textbf{blue}} lines).

In this work, we introduce \textbf{{Sw}}apping \textbf{{A}}ssignments for \textbf{{Ro}}bust contrastive learning (\textbf{\modelname}), a simple and efficient alternative that leverages cluster-wise supervision to design semi-targeted adversarial attacks that missguide the contrastive loss towards true negatives, \ie accounting for examples that have pseudo-label assignments different than the anchor class. Specifically, our method updates the gradient direction to obtain adversarial perturbations that maximize the loss for pairs with same pseudo-labels (Figure~\ref{fig:sketch} \textcolor{blue}{\textbf{blue}} lines), but also minimize the loss of negative pairs (Figure~\ref{fig:sketch} \textcolor{red}{\textbf{red}} lines), leading to pseudo-targeted attacks.

\noindent \textbf{Contributions} The contributions of our work are summarized as follows: 

\noindent \textbf{(1)} We introduce \modelname, a self-supervised contrastive learning method that jointly considers adversarial robustness and cluster-wise semantic information observed in the data. 

\noindent \textbf{(2)} In contrast to prior work on adversarial contrastive learning, we utilize the pseudo-labels induced by clustering as targets to explicitly guide the gradient direction towards negatives. 

\noindent \textbf{(3)} We verify the effectiveness of the proposed method and show that \modelname improves downstream performance (accuracy), robustness on unseen black-box and white-box attack types, and transfer learning performance.
\section{Related Work}\label{sec:related_work}
\paragraph{Adversarial Training}
There has been substantial research on increasing the robustness of deep neural networks against adversarial attacks. For example, \citet{Goodfellow2015ExplainingAH} propose the Fast Gradient Sign Method (FGSM), a white-box attack that generates adversarial perturbations by utilizing the model gradients to design small distortions, which when added to the original image will lead to maximizing the model loss. Training with such adversarial perturbations has been shown to improve model robustness. Subsequent work proposes iterative variants, \eg extensions such as the Basic Iterative (BIM)~\cite{Kurakin2017AdversarialEI} and Projected Gradient Descent (PGD)~\cite{madry2017towards} methods. Besides the gradient-based cross-entropy attacks, TRADES~\cite{zhang2019theoretically} empirically analyzes the trade-off between robustness and accuracy and introduces a defense mechanism that utilizes a KL-divergence loss between a clean example and its adversarial counterpart for learning a more robust latent feature space. \citet{ilyas2019adversarial} hypothesize that the adversarial vulnerability of neural networks is a direct result of their sensitivity to well-generalizing features that are incomprehensible to humans. \citet{Schwinn2021ExploringMO} discover that cross-entropy attacks fail against models with large logits, and propose to add logit noise and enforce scale invariance on the loss to mitigate this limitation and encourage the model to design diverse attack targets. All above methods are originally designed for supervised learning tasks.

\paragraph{Contrastive Learning}\label{rel_work:cl}
Instance-wise contrastive learning methods~\cite{ye2019unsupervised,chen2020simclr, bachman2019learning, he2020momentum, oord2018representation, caron2021emerging} learn an embedding space where different views of the same instance lie close to each other and views of dissimilar instances are pushed further away. For example, SimCLR~\cite{chen2020simclr} incorporates various data augmentation techniques to generate positive examples for contrastive training. MoCo~\cite{he2020momentum} utilizes a dynamic memory bank and a moving average encoder for calculating the contrastive loss. \citet{he2020momentum} experiment with the number of negative examples and found that using more negatives results in learning better feature representations.
However, due to only considering different views of the same instance as positive samples, instance-wise contrastive learning can not learn representations that capture the global class-level semantic structure of the training data. To alleviate this problem, clustering-based or prototypical contrastive learning~\cite{sharma2020clustering,Li2020PCL,goyal2021self} and supervised contrastive learning~\cite{khosla2020supervised} methods have been proposed. 
For example, \citet{Li2020PCL} utilize a cluster-based approach that brings contrastive representations closer to their assigned feature prototypes. 
\citet{Caron2020Cluster} propose SwAV, an image representation technique that instead of comparing representations of different image views, follows a swapped prediction strategy to predict the code of one view from another view of the same image by clustering image representations with a computationally efficient online clustering approach. Similar swapping strategies have been utilized in cross-modal retrieval~\cite{kim2021swamp}.
SEER~\cite{goyal2021self} is trained on large-scale unconstrained and uncurated image collections by improving the scalability of SwAV in terms of GPU memory consumption and training speed. \citet{khosla2020supervised} extend contrastive learning to a supervised setting that utilizes label information in positive-negative subset creation, \ie images of the same class label are considered as positive examples. 
All aforementioned instance-wise and cluster-based contrastive learning techniques lack robustness against adversarial examples. Here, we briefly review works that aim to combine adversarial training with contrastive learning.

\paragraph{Adversarial Contrastive Training}\label{rel_work:at}
A number of works utilize adversarially constructed samples in contrastive learning to increase the robustness of the learned feature representations ~\cite{bui2021understanding, Kim2020AdversarialCL,Jiang2020PretrainingACL,Chen_2020_CVPR, mcdermott2021adversarial, chen2020self}. RoCL~\cite{Kim2020AdversarialCL} introduces small perturbations to generate adversarial samples, leading to a more robust contrastive model. \citet{Jiang2020PretrainingACL} experiment with different combinations of perturbations while generating positive and negative examples for calculating contrastive loss and found that using both standard data augmentations and adversarial samples increases robustness in a self-supervised pre-training procedure. 
Both works do not consider clustered prototypes during the contrastive loss computation or the adversarial sample generation process. \citet{Chen_2020_CVPR} analyze robustness gains after applying adversarial training in both self-supervised pre-training and supervised fine-tuning, and find that adversarial pre-training contributes the most in robustness improvements. \citet{bui2021understanding} propose a minimax local-global algorithm to generate adversarial samples that improves the performance of SimCLR~\cite{chen2020simclr}. To the best of our knowledge, none of the previous works considered designing more targeted contrastive adversarial examples by leveraging useful cluster structures observed in the data.

\begin{figure*}[t!]
    \centering
    \includegraphics[width=\linewidth]{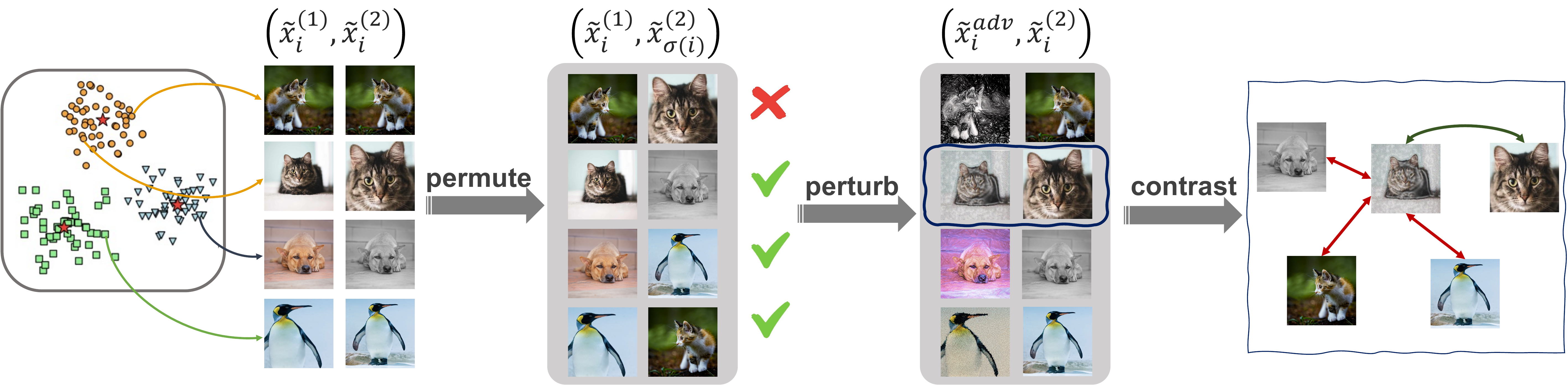}
    \caption{{Pictorial overview of our proposed adversarial contrastive learning method \modelname. First, examples are pseudo-labeled via clustering and positive pairs are created by applying data augmentations to each example. Then, pairs are permuted to produce targets for the adversarial attack, that is guided by the pseudo-labels based on whether the permuted pair contains images from the same cluster or not. This process generates adversarial examples that are finally matched with their corresponding original pair examples to compute the contrastive loss. Inspired by targeted attacks, \modelname minimizes the loss for negative pairs and maximizes the loss for positive pairs. In contrast prior works only maximize the self-supervised contrastive loss between positive examples.
    }}
    \label{fig:overview}
\end{figure*}

\section{Proposed Method}\label{sec:proposed_method}
\paragraph{Problem Formulation}
Given a set of unlabeled instances $\mathcal{X}=\{x_1,x_2,...,x_n\}$, the goal is to learn a feature encoder $f_{\theta}(\cdot)\colon \mathcal{X} \to \mathbb{R}^d$ that maps data points to a $d$-dimensional embedding space and can later on be used in downstream tasks. Contrastive learning methods learn such feature representations by minimizing the distance between similar data points (positive pairs) and maximizing the distance between dissimilar data points (negative pairs). 
The contrastive loss is typically calculated via a normalized temperature-scaled cross-entropy loss (NT-XENT) as shown in Eq. (\ref{eq:ntxent_loss}). 
\begin{align}
\mathcal{L}_{CL}(x_{i}, x_k) \,{=}\, -\log \frac{ \exp \big( s(x_i,x_k) / \tau \big) }
{ \sum\limits_{{x}_{j {\neq} i,k} \in \mathcal{X}} \exp \big(  s(x_i, x_j) / \tau \big) },
\label{eq:ntxent_loss}
\end{align}
where $s(x_i,x_j)  \,{=}\, f_{\theta}(x_i)^{\top}f_{\theta}(x_j) / \lVert f_{\theta}(x_i) \rVert \lVert f_{\theta}(x_j) \rVert$ is the cosine similarity between two  latent representations, $\tau$ is a temperature parameter, and $(x_i,x_k)$ is the positive pair while all other instances ${x}_{j {\neq} i,k} \in \mathcal{X}$ are considered negative pairs. The final contrastive loss is averaged across all positive pairs in a batch of examples.

It has been shown, however, that such formulations of contrastive learning are vulnerable against adversarial attacks~\cite{Kim2020AdversarialCL}.
Subsequently, a few adversarial self-supervised methods have been proposed that mainly apply adversarial training to contrastive pretraining such that the feature encoder $f_{\theta}$ learns robust data representations~\cite{Kim2020AdversarialCL,Jiang2020PretrainingACL}.

\paragraph{Supervised \& Self-supervised Adversarial Training}
In a supervised setting, adversarial training methods search for the worst-case instance perturbation that maximizes the supervised loss. To ensure that the semantic content of the instance does not alter, the perturbation space is constrained within a certain radius (\eg $\ell_\infty$ norm-ball of radius $\epsilon$). The training objective is the following mini-max optimization problem:
\begin{equation}
\argmin_{\theta}\mathbb{E}_{{(x_i,y_i)}\sim \mathcal{D}} \left[ \max_{\| \delta_i\|_\infty\leq\epsilon} \mathcal{L}_{CE}(x_i+\delta_i, y_i; \theta) \right],
\label{eq: AT}
\end{equation}
where $f_{\theta}\colon \mathcal{X} \to \mathcal{Y}$ denotes the supervised learning model with $\theta$ model parameters, $\mathcal{L}_{CE}$ is the supervised training objective, \eg cross-entropy loss, and $(x_i, y_i) \in \mathcal{D}$ denotes a training sample with feature $x_i \in \mathcal{X}$ and corresponding label $y_i \in \mathcal{Y}$. Here, $\delta_i$ is the adversarial perturbation and subsequently $x_i+\delta_i$ is an adversarial example. There have been several gradient-based methods proposed to solve the inner maximization problem, essentially adjusting $\delta_i$ in the direction that maximizes the gradient $\nabla_{\delta_i} \mathcal{L}_{CE}(f_\theta(x_i + \delta_i),y_i)$~\cite{Goodfellow2015ExplainingAH,Kurakin2017AdversarialEI,madry2017towards,zhang2019theoretically,croce2020reliable,Schwinn2021ExploringMO}. For example, the Projected Gradient Descent (PGD)~\cite{madry2017towards} iterates over gradient steps and adjusts $\delta$ accordingly:
\begin{equation}
\delta_i^{t+1} := \Pi \Big[ \delta_i^{t} + \eta \texttt{sign}\left(\nabla_{\delta_i} \mathcal{L}_{CE}\left(f_\theta(x_i+\delta_i^{t}), y_i\right) \right) \Big],
\end{equation}
where $t$ denotes the number of iterations, $\eta$ is the step size, $\texttt{sign}(\cdot)$ is the vector sign (direction), and $\Pi$ denotes the norm-ball projection of interest, \eg clipping $\delta_i$ to lie within an $\epsilon$-range $\Delta:=\{\delta\colon \|\delta_i\|_\infty \leq \epsilon\}$.
Notice that the above formulation requires a class label $y \in \mathcal{Y}$ to compute the supervised loss when crafting adversarial attacks and hence is inapplicable to self-supervised settings. Also, depending on the choice of $y$, the attack can be untargeted or targeted. In targeted adversarial attacks the goal is to direct the perturbation towards a particular class of interest, other than the true class, \eg confusing the model to misclassify a ``cat'' example as ``dog'', by using as target $y$ the class label for ``dog'' instead of the model prediction.

To adapt adversarial training in self-supervised settings, recent work replaces the supervised loss with the contrastive NT-XENT loss in Eq.(\ref{eq:ntxent_loss}), \ie generating adversarial views of an instance that confuse the model w.r.t. the instance identify (semantic class). However, this straightforward application of adversarial training can be considered as an untargeted attack that lacks the discriminative ability that observed data cluster structure (class) information can provide. In the following section, we present our proposed adversarial contrastive training approach.

\subsection{\modelname Description}
Our proposed adversarial contrastive method consists of three steps. At first, we assign each training sample into clusters via clustering. The choice of the clustering method is orthogonal to our approach. Then, we generate adversarial samples by perturbing the instances in each example pair, and guide the gradient for each example pair towards a corrected direction, based on whether the cluster assignments (pseudo-label) for the two instances are similar (and hence the two examples are most likely to encode a positive pair) or not (and this can be indeed considered as a negative pair). Afterward, we calculate the contrastive loss and update the encoder model. Each step is discussed in detail. 

\paragraph{Cluster Assignments}\label{sec:clusterassign}
We utilize clustering to obtain cluster centroids and assign a pseudo-label to each training sample. For simplicity, we apply $k$-means, however, we note that any clustering method can be used, including online clustering variants~\cite{asano2019self,caron2018deep,Caron2020Cluster}. 
More formally, let $\mathcal{Z} \,{=}\, \{z_1,z_2,...,z_n\}$ be the intermediate latent representations obtained from feature encoder $f_\theta$ for unlabeled dataset $\mathcal{X}$, and $K$ is the empirically-set number of clusters. A set of cluster centroids $\mathcal{C} \,{=}\, \{c_j\}_{j=1}^{K}$ can be computed by applying $k$-means clustering on $\mathcal{Z}$. Next, we assign pseudo-label $\hat{y}_{x_i}$ to each example $x_i \in\mathcal{X}$ based on the distance from each cluster centroid, \ie $\hat{y}_{x_i}  \,{=}\, \argmin_{j}{{(z_i - c_j)}^2}$.

\paragraph{Adversarial Sample Generation}
Let $\mathcal{X}_B \,{=}\, \{x_i\}^{B}_{i=1} \subset \mathcal{X}$ sampled batch of size $B$. To construct positive pairs, an example is augmented twice by applying two transformations $t_1, t_2 \in \mathcal{\Tau}$ that create two different views $\tilde{x}_i^{(1)} \,{=}\, t_1(x_{i})$ and $\tilde{x}_i^{(2)} \,{=}\, t_2(x_{i})$ of the same instance, producing a batch of positive pairs $\{(\tilde{x}_1^{(1)}, \tilde{x}_1^{(2)}), \ldots, (\tilde{x}_i^{(1)}, \tilde{x}_i^{(2)}), \ldots, (\tilde{x}_B^{(1)}, \tilde{x}_B^{(2)})\}$. Without  loss  of  generality, a permutation induced from shuffling the second instance in each pair is denoted as: 
\begin{equation}\label{eq:permutation}
     \{(\tilde{x}_1^{(1)}, \tilde{x}_{\sigma(1)}^{(2)}), \ldots, (\tilde{x}_i^{(1)}, \tilde{x}_{\sigma(i)}^{(2)}), \ldots, (\tilde{x}_B^{(1)}, \tilde{x}_{\sigma(B)}^{(2)})\},
\end{equation}
where $\sigma(\cdot)\colon \{1, \ldots, B\} \to \{1, \ldots, B\}$ is a permutation.
Based on the pseudo-labels created from clustering (Section \ref{sec:clusterassign}), we can assign an indicator variable to each pair in the batch, according to whether both instances belong to the same cluster, \ie
\begin{align}\label{eq:indicator}
 \mathcal{I}{\left(\tilde{x}_i^{(1)}, \tilde{x}_{\sigma(i)}^{(2)}\right)} \,{=}\, \left[ \mathbbm{1}\left(\hat{y}_{\tilde{x}_i^{(1)}} \,{=}\, \hat{y}_{\tilde{x}_{\sigma(i)}^{(2)}}\right)
 - \mathbbm{1}\left(\hat{y}_{\tilde{x}_i^{(1)}} \,{\neq}\, \hat{y}_{\tilde{x}_{\sigma(i)}^{(2)}}\right) \right],
\end{align}
where $\hat{y}_{\tilde{x}_i^{(1)}}$ and $\hat{y}_{\tilde{x}_{\sigma(i)}^{(2)}}$ are pseudo-labels from clustering.

In short, we assign $\mathcal{I}{\left(\tilde{x}_i^{(1)}, \tilde{x}_{\sigma(i)}^{(2)}\right)} = 1$ to a pair if both of the examples in that pair belongs to the same cluster (positive pair) and $\mathcal{I}{\left(\tilde{x}_i^{(1)}, \tilde{x}_{\sigma(i)}^{(2)}\right)}  = -1$ if they belong to different clusters (negative pair).
We utilize this indicator to better guide the \texttt{sign} of the NT-XENT gradient w.r.t. the corresponding example pair. In essence, we maximize the contrastive loss w.r.t. positive pairs, and minimize the contrastive loss w.r.t. negative pairs, as shown in Eq. (\ref{eq:gradient_update}):
\begin{align}
\delta_i^{t+1} := \Pi \left[
\delta_i^{t} + \eta~ {\beta}_{x_i} \texttt{sign}
\left(
\nabla_{\delta_i} 
\mathcal{L}_{CL}\left(
f_{\theta}\left(\tilde{x_i}^{(1)}+\delta_i^{t}\right),~
f_{\theta}\left(\tilde{x}_{\sigma(i)}^{(2)}\right)
\right)
\right)
\right],
\label{eq:gradient_update}
\end{align}
where ${\beta}_{x_i} \,{=}\, \mathcal{I}{\left(\tilde{x}_i^{(1)}, \tilde{x}_{\sigma(i)}^{(2)}\right)}$ for brevity.
Notice that $f_{\theta}\left(\tilde{x}_{\sigma(i)}^{(2)}\right)$, in combination with ${\beta}_{x_i}$, acts as a target for the adversarial attack.
Applying these pertubations to each example in the batch creates adversarial examples $x^{adv}_i \,{=}\, \left(\tilde{x}_i^{(1)}+ \delta_{i}^{(1)}, \tilde{x}_{\sigma(i)}^{(2)}\right)$ that aim to maximize the similarity when both examples in a pair belong to different clusters. 

\paragraph{Contrastive Training}
Finally, pairs are reordered in their original configuration $\mathcal{X}^{adv} \,{=}\, \{(\tilde{x}_i^{(1)} + \delta_{i}^{(1)},~ \tilde{x}_i^{(2)})\}_{i=1}^{B}$, with one instance in each pair being adversarially attacked, and the model is trained with the contrastive NT-XENT loss computed as in Eq.(\ref{eq:ntxent_loss}).
Figure~\ref{fig:overview} presents an overview and Algorithm~\ref{alg:pseudo} provides the pseudocode of the proposed method.
\begin{algorithm}[!t]
\caption{\label{alg:pseudo} Pseudocode for \modelname}
\begin{algorithmic}
    \STATE \textbf{Input:} Unlabeled dataset $\mathcal{X}$, batch size $B$, encoder $f_{\theta}$, transformations $\mathcal{\Tau}$
    \FOR{batch $\mathcal{X}_B \,{=}\, \{(x_{i})\}_{i=1}^{B} \subset \mathcal{X}$}
    \STATE $\texttt{loss} := 0$, $\mathcal{S} \,{=}\, \emptyset$
        \STATE \textbf{for} $i = 1, \ldots, B$ \textbf{do}
         $~~$\textcolor{gray}{\# Augment $x_i$ with transformations sampled from $\mathcal{\Tau}$}
        \STATE $~~$ $\tilde{x}_i^{(1)} \,{=}\, t_1(x_{i})$ and $\tilde{x}_i^{(2)} \,{=}\, t_2(x_{i})$
        \STATE $~~$ $\mathcal{S} \,{=}\, \mathcal{S} \cup \left\{\left(\tilde{x}_i^{(1)}, \tilde{x}_i^{(2)}\right)\right\}$~\textcolor{gray}{\# collect pairs}
        \STATE \textbf{end for}
    \STATE $\sigma(\mathcal{S}) \,{=}\, \left\{ \left(\tilde{x}_i^{(1)}, \tilde{x}_{\sigma(i)}^{(2)}\right)\right\}_{i=1}^{B}$ 
    $~~$ \textcolor{gray}{\# Permute instances}
    \STATE \textbf{for} $i = 1, \ldots, B$ \textbf{do}
    \STATE $~~$ ${\beta}_{x_i} \,{=}\, \left[ \mathbbm{1}\left(\hat{y}_{\tilde{x}_i^{(1)}} \,{=}\, \hat{y}_{\tilde{x}_{\sigma(i)}^{(2)}}\right) - \mathbbm{1}\left(\hat{y}_{\tilde{x}_i^{(1)}} \,{\neq}\,\hat{y}_{\tilde{x}_{\sigma(i)}^{(2)}}\right) \right]$
      $~~$ \textcolor{gray}{\# Assign indicator to each pair via Eq.(\ref{eq:indicator})}
    \STATE $~~$ \textbf{for} $t = 1, \ldots, T$ \textbf{do}
    \STATE $~~~~~~~$ $\delta_i^{t+1} := \Pi \left[\delta_i^{t} + \eta~ {\beta}_{x_i} \texttt{sign}\left(\nabla_{\delta_i}\mathcal{L}_{CL}\left(f_{\theta}\left(\tilde{x_i}^{(1)}+\delta_i^{t}\right),~f_{\theta}\left(\tilde{x}_{\sigma(i)}^{(2)}\right)\right)\right)\right],$
    \STATE $~~~~~~$ \textcolor{gray}{\# Compute adversarial perturbations via Eq.(\ref{eq:gradient_update})}
    \STATE $~~$ \textbf{end for}
    \STATE $~~$ $x^{adv}_i \,{=}\, \left(\tilde{x}_i^{(1)}+ \delta_{i}^{(1)}, ~\tilde{x}_{i}^{(2)}\right)$ 
    $~~$ \textcolor{gray}{\# Apply adversarial permutation to original pair}
    \STATE $~~$ $\mathcal{S} \,{=}\, \mathcal{S}\setminus \left\{\left(\tilde{x}_i^{(1)}, ~\tilde{x}_i^{(2)}\right)\right\} \cup x^{adv}_i$~\textcolor{gray}{\# Update $\mathcal{S}$}
    \STATE \textbf{end for}
    \STATE \texttt{loss} += $ \sum\limits_{i\in B} \left[ - \log \frac{ \exp \big( s\left(\tilde{x}_i^{(1)} + \delta_{i}^{(1)}, \tilde{x}_i^{(2)}\right)/ \tau \big) }
{ \sum\limits_{{x}_{j {\neq} i} \in \mathcal{X}} \exp \big( s\left(\tilde{x}_i^{(1)} + \delta_{i}^{(1)}, \tilde{x}_i^{(2)}\right) / \tau \big)} \right]$ 
$~~$ \textcolor{gray}{\# calculate contrastive loss}
    \STATE \textbf{end for}
    \STATE update model parameters $\theta$ to minimize the loss
    \ENDFOR
\end{algorithmic}
\end{algorithm}

\section{Experimental Results}\label{sec:experiment}
We evaluate \modelname on two benchmark datasets comparing against existing self-supervised and supervised adversarial learning methods, and reporting results on black-box and white-box targeted and untargeted attacks. Similarly to \citet{Kim2020AdversarialCL}, we consider a linear evaluation and a robust-linear evaluation (\textbf{r-LE}) setting, where we compare the classification accuracy on clean and adversarial examples, respectively. 

\paragraph{Experimental Setup}
For all experiments, we use a ResNet-18~\cite{he2016deep} as the backbone encoder. All models, including baselines are trained with $\ell_\infty$ Projected Gradient Descent (PGD) with $\epsilon = 8/255$. As for the linear evaluation, we follow a similar setup as RoCL~\cite{Kim2020AdversarialCL}, where we first train an unsupervised model and then train a linear layer to measure the accuracy. All reported results are averaged over multiple trials, and hyper-parameters are provided in the supplementary material. 
To ensure the efficacy of the clustering process, we rely on instance-wise adversarial attacks for the first 100 epochs and then utilize the proposed cluster-wise adversarial example generation. This is to guarantee that the encoder has learned useful representations before initiating clustering. 

\paragraph{Datasets \& Baselines}
We conduct experiments on the following datasets:

\noindent \texttt{\textbf{CIFAR-10}}~\citep{cifar10_100}: The CIFAR-10 dataset includes $60,000$ images of size $32 \times 32$, originating from $10$ classes, with $6,000$ images per class. The training and test sets consist of $50,000$ and $10,000$ images, respectively.

\noindent \textbf{\texttt{CIFAR-100}}~\citep{cifar10_100} contains images similar to \texttt{{CIFAR-10}} but there exist $100$ classes and $600$ images per class. Train-to-test ratio remains the same as \texttt{{CIFAR-10}}.

Moreover, we compare \modelname against six types of baseline methods: (1) Supervised models trained with cross-entropy (\textbf{CE}), (2) traditional adversarial training methods such as \textbf{AT}~\cite{madry2017towards} and \textbf{TRADES}~\cite{zhang2019theoretically}, (3) a vanilla contrastive learning method, \textbf{SimCLR}~\cite{chen2020simclr}, (4) supervised contrastive learning (\textbf{SCL}) ~\cite{khosla2020supervised}, (5) \textbf{SwAV}~\cite{Caron2020Cluster}, a contrastive method that utilizes clustering, and (6) adversarial contrastive learning methods such as \textbf{RoCL}~\cite{Kim2020AdversarialCL} and \textbf{ACL}~\cite{Jiang2020PretrainingACL}.

\begin{table}[t!]
\caption{Comparison against PGD white-box attacks on
\texttt{CIFAR-10}. Results marked with \textsuperscript{\textdagger} are reported from ~\citet{Kim2020AdversarialCL}. \textsc{CE} is supervised cross-entropy. $A_{nat}$ denotes the accuracy of clean images. $\ell_\infty$ is the \textit{seen} adversarial attack while $\ell_1$  and $\ell_2$ are \textit{unseen} attacks. Colored fonts indicate best performance in each category: supervised (\textcolor{violet}{\textbf{violet}}), self-supervised (\textcolor{teal}{\textbf{teal}}), adversarial self-supervised on linear evaluation (\textcolor{black}{\textbf{black}}), and robust linear evaluation (\textcolor{purple}{\textbf{purple}}). 
}
  \label{tab:accuracy_white_box}
  \resizebox{\columnwidth}{!}{
    \centering
{
  \begin{tabular}{ccccccccc} 
    \toprule
    \multirow{3}{*}{\makecell{\\\\Train type}}&\multirow{3}{*}{\makecell{\\\\Method}} 
    {}&{}& \multicolumn{2}{c}{\emph{seen}} & \multicolumn{4}{c}{\emph{unseen}} \\
    {}&{}&{}&\multicolumn{2}{c}{$\ell_{\infty}$}&\multicolumn{2}{c}{$\ell_{2}$}&\multicolumn{2}{c}{$\ell_{1}$}\\
    \cmidrule(r){4-5} \cmidrule(r){6-7} \cmidrule(r){8-9} 
    {}&{}&{$A_{nat}$}& $\epsilon$ 8/255 & 16/255 &0.25&0.5&7.84&12   \\
    \midrule \midrule
    \multirow{6}{*}{Supervised}& \textsc{CE}\textsuperscript{\textdagger}&92.82&0.00&0.00&20.77&12.96&28.47&15.56\\
    {}&{AT}\textsuperscript{\textdagger} ~\cite{madry2017towards}&81.63&44.50&14.47&\textcolor{violet}{\textbf{72.26}}&\textcolor{violet}{\textbf{59.26}}&\textcolor{violet}{\textbf{66.74}}&\textcolor{violet}{\textbf{55.74}}\\
    {}&TRADES\textsuperscript{\textdagger}~\cite{zhang2019theoretically}&77.03&\textcolor{violet}{\textbf{48.01}}&\textcolor{violet}{\textbf{22.55}}&68.07&57.93&62.93&53.79\\
    {}&SCL\textsuperscript{\textdagger}~\cite{khosla2020supervised}&\textcolor{violet}{\textbf{94.05}}&0.08&0.00&22.17&10.29&38.87&22.58\\
    \midrule \midrule
    \multirow{2}{*}{\makecell{Self-\\supervised}} & SimCLR\textsuperscript{\textdagger}~\cite{chen2020simclr}&\textcolor{teal}{\textbf{91.25}}&0.63&0.08&15.3&2.08&\textcolor{teal}{\textbf{41.49}} &25.76\\
    {}&SwAV~\cite{Caron2020Cluster}& 70.60 & \textcolor{teal}{\textbf{22.88}} & \textcolor{teal}{\textbf{18.46}} & \textcolor{teal}{\textbf{35.54}} & \textcolor{teal}{\textbf{35.53}} & 35.54 & \textcolor{teal}{\textbf{35.54}} \\
    \midrule
    \multirow{3}{*}{\makecell{Adversarial\\ Self-supervised}}&RoCL\textsuperscript{\textdagger}~\cite{Kim2020AdversarialCL}&83.71&40.27&9.55&66.39&63.82& {79.21} & {76.17}\\
    {}&ACL~\cite{Jiang2020PretrainingACL}& 80.91 & \textcolor{black}{\textbf{41.39}} & \textcolor{black}{\textbf{12.95}} & \textcolor{black}{\textbf{75.00}} & \textcolor{black}{\textbf{75.45}} & 79.32 & 79.30    \\
    {}&{\textbf{\modelname (Ours)}}& \textcolor{black}{\textbf{87.38}} & 37.33 & 9.13 & 59.69 & 59.10 & \textcolor{black}{\textbf{84.06}} & \textcolor{black}{\textbf{83.69}} \\
    \midrule 
    \multirow{2}{*}{\makecell{Adversarial (Robust Eval.)\\ Self-supervised}}
    &RoCL+{rLE}\textsuperscript{\textdagger} & 80.43 & \textcolor{purple}{\textbf{47.69}} &\textcolor{purple}{\textbf{15.53}}&\textcolor{purple}{\textbf{68.30}}&\textcolor{purple}{\textbf{66.19}}&77.31&75.05\\
    {}&{\textbf{\modelname (Ours) + rLE}}& \textcolor{purple}{\textbf{86.89}} & 41.58 & 10.70 & 62.20 & 62.66 &\textcolor{purple}{\textbf{84.09}} & \textcolor{purple}{\textbf{83.79}} \\
    \bottomrule
  \end{tabular}
  }
  }
\end{table}
\begin{table}[t]
\caption{Comparison against PGD white-box attacks on
\texttt{CIFAR-100}. $A_{nat}$ denotes the accuracy of clean images. $\ell_\infty$ is the \textit{seen} adversarial attack while $\ell_1$  and $\ell_2$ are \textit{unseen} attacks.}
  \label{tab:cifar100}
  \centering
  \resizebox{.75\textwidth}{!}{
{
  \begin{tabular}{cccccccc} 
    \toprule
    \multirow{3}{*}{\makecell{\\\\Method}} 
    &{}& \multicolumn{2}{c}{\emph{seen}} & \multicolumn{4}{c}{\emph{unseen}} \\
    \cmidrule(r){3-4} \cmidrule(r){5-8}  
    {}&{}&\multicolumn{2}{c}{$\ell_{\infty}$}&\multicolumn{2}{c}{$\ell_{2}$}&\multicolumn{2}{c}{$\ell_{1}$}\\
    \cmidrule(r){3-4} \cmidrule(r){5-6} \cmidrule(r){7-8} 
    {}&{$A_{nat}$}& $\epsilon$ 8/255 & 16/255 &0.25&0.5&7.84&12   \\
    \midrule \midrule
    ACL~\cite{Jiang2020PretrainingACL} & 42.03 & \textbf{22.97} & \textbf{9.53} & \textbf{38.37} & \textbf{38.67} & 40.92 & 40.88 \\
    RoCL~\cite{Kim2020AdversarialCL} & 50.04 & 20.26 & 5.93 & 36.16 & 36.09 & \textbf{52.44} & 52.08 \\
    \textbf{\modelname (Ours)} & \textbf{55.12} &  20.19 & 5.96 & 36.37 & 36.11 & \textbf{52.44} & \textbf{52.18} \\
    \bottomrule
  \end{tabular}
  }
  }
\end{table}
\subsection{White Box Attacks}
We first evaluate the robustness of the learned representations against seen and unseen attacks. 
In Table~\ref{tab:accuracy_white_box}, we report results on \texttt{CIFAR-10}, comparing methods against Projected Gradient Descent (PGD) white-box attacks with linear evaluation and robust linear evaluation (rLE).~
Table~\ref{tab:cifar100} presents results for all adversarial self-supervised methods on \texttt{CIFAR-100}.

Traditional contrastive learning methods (SimCLR and SwAV) are vulnerable to adversarial attacks. Considering cluster information (SwAV) improves performance on seen $\ell_\infty$ attacks and unseen $\ell_2$ attacks but occurs considerable losses in accuracy on clean images $A_{nat}$. Among adversarial self-supervised approaches, \modelname achieves much
higher clean accuracy, substantially higher robust accuracy against unseen $\ell_1$ target attacks, and comparable performance with RoCL on $\ell_\infty$ and $\ell_2$ attacks. ACL outperforms both on $\ell_\infty$ and $\ell_2$; this method benefits from additional batch normalization parameters and balancing two contrastive loss terms, a standard contrastive and an adversarial contrastive loss. 

In addition to PGD, we also experiment with several additional state-of-the-art white-box attacks BIM~\cite{Kurakin2017AdversarialEI}, FGSM~\cite{Goodfellow2015ExplainingAH}, Jitter~\cite{Schwinn2021ExploringMO}. \modelname outperforms baselines across all untargeted and targeted white-box attacks, showing consistent robust accuracy gains (Table \ref{tab:white_box}). In particular, we observe approximately $3-4\%$ relative gains on PDG variants such as BIM and FGSM w.r.t. the next best method. Additionally, we find that \modelname is performing $4-9\%$ better compared to baselines on Jitter attacks, an adversarial attack method that incorporates scale invariance and encourages diverse attack targets with smaller perturbations~\cite{Schwinn2021ExploringMO}. These results make \modelname more appealing in practice and suggest that our approach to enforcing adversarial perturbations, that consider both positive and negative pairs as well as semantic cluster information, ensures robustness against a diverse set of attack types.
\begin{table}[t!]
\caption{Performance of \modelname against white-box attacks (BIM, FGSM, Jitter) on the \texttt{CIFAR-10} dataset. Each column denotes the white-box attack used to generate adversarial samples with $\epsilon=$8/255. Each row shows the performance of the target models against the white-box attacks.}
  \label{tab:white_box}
  \resizebox{\textwidth}{!}{
    \centering
{
  \begin{tabular}{lcccccc}
  \toprule
    \multirow{2}{*}{\backslashbox{Method}{Attack}} & \multicolumn{2}{c}{BIM~\cite{Kurakin2017AdversarialEI}}                                       & \multicolumn{2}{c}{FGSM~\cite{Goodfellow2015ExplainingAH}}                                      & \multicolumn{2}{c}{Jitter~\cite{Schwinn2021ExploringMO}}                                    \\
     \cmidrule(r){2-3}\cmidrule(r){4-5}\cmidrule(r){6-7}
     & \multicolumn{1}{c}{Untargeted} & \multicolumn{1}{c}{Targeted} & \multicolumn{1}{c}{Untargeted} & \multicolumn{1}{c}{Targeted} & \multicolumn{1}{c}{Untargeted} & \multicolumn{1}{c}{Targeted} \\
     \midrule\midrule
RoCL & 37.07 & 62.92 & 57.27 & 67.06 & 45.41 & 63.67 \\
ACL  & {42.35}  & {65.74} & 49.95 & {63.02} & 54.24 & 61.55   \\                    
\textbf{\modelname (Ours)} & $~~~~~~$\textbf{45.78}\color{green}$^{\bf\uparrow3.43}$ & $~~~~~~$\textbf{68.70}\color{green}$^{\bf\uparrow2.96}$ & $~~~~~~$\textbf{61.34}\color{green}$^{\bf\uparrow4.07}$ & $~~~~~~$\textbf{67.83}\color{green}$^{\bf\uparrow0.77}$ & $~~~~~~$\textbf{57.75}\color{green}$^{\bf\uparrow3.51}$ & $~~~~~~$\textbf{72.57}\color{green}$^{\bf\uparrow8.90}$ \\
\bottomrule
\end{tabular}
  }
  }
\end{table}

\begin{table*}[t!]
\caption{Performance comparison against black-box attacks on the \texttt{CIFAR-10} dataset. Columns denote black-box $\ell_{\infty}$ attacks and rows show model performance (models trained with $\ell_{\infty}$).}
  \label{table:blackbox-table}
  \centering
  \resizebox{\textwidth}{!}{
    \centering
  \begin{tabular}{lllllllll}
    \toprule
    \multirow{2}{*}{\backslashbox{Target}{Source}}&\multicolumn{4}{c}{ $\epsilon=$8/255} &\multicolumn{4}{c}{ $\epsilon=$16/255}  \\
    \cmidrule(r){2-5} \cmidrule(r){6-9}
    & AT& TRADES&  RoCL & \modelname 
    & AT& TRADES &  RoCL & \modelname\\
    \midrule
     AT~\cite{madry2017towards}&-& 41.67$\pm$0.06& 43.06$\pm$0.08&43.53$\pm$0.05&-& 36.97$\pm$0.08&40.81$\pm$0.14&41.89$\pm$0.06\\
     TRADES~\cite{zhang2019theoretically}& 82.29$\pm$0.04&-& {74.43$\pm$0.04}&{73.31$\pm$0.10}& 80.10$\pm$0.09&-&{55.88$\pm$0.11}&{55.55$\pm$0.07}\\
     RoCL~\cite{Kim2020AdversarialCL}&  82.03$\pm$0.07 &  71.39$\pm$0.02&-&\textcolor{blue}{61.41$\pm$0.08}& 79.11$\pm$0.08& \textbf{46.09$\pm$0.10}&-&\textcolor{blue}{28.86$\pm$0.15}\\
     ACL~\cite{Jiang2020PretrainingACL}&  80.28$\pm$0.05 &  69.62$\pm$0.04 & 63.53$\pm$0.04 & 63.06$\pm$0.10& 77.93$\pm$0.13& 45.56$\pm$0.03& 32.48$\pm$0.07 &33.33$\pm$0.14\\
     \textbf{\modelname (Ours)} &  \textbf{84.99$\pm$0.12} &  \textbf{74.38$\pm$0.08} & \textcolor{blue}{64.27$\pm$0.12}&-&  \textbf{81.80$\pm$0.23} &   45.40$\pm$0.09&\textcolor{blue}{25.61$\pm$0.15}&-\\
    \bottomrule
  \end{tabular}}
\end{table*}

\subsection{Black Box Attacks}
We also verify the performance of \modelname against black-box attacks. We generate adversarial examples using black-box $\ell_\infty$ attacks such as AT~\cite{madry2017towards}) and TRADES~\cite{zhang2019theoretically}. Table~\ref{table:blackbox-table} shows that \modelname is able to better defend against TRADES and AT attacks.
We additionally compare \modelname as a black-box adversarial attack on TRADES, AT, RoCL and ACL. Results show that \modelname attacks are stronger or comparable to baselines. An interesting observation is that as the perturbation radius $\epsilon$ becomes smaller, \modelname as an attack is stronger than RoCL, as \modelname's performance on RoCL attacks is  $64.27\%$, compared to RoCL's performance on \modelname attacks that drops to $61.41\%$. Similarly, we also find that ACL and TRADES are better equipped against RoCL attacks than \modelname attacks.

\subsection{Transfer Learning}
We evaluate performance on a transfer learning setup and check whether the learned representations can be transferred across datasets. Specifically, we pretrain an encoder model on \texttt{CIFAR-10} and then finetune a randomly-initialized linear classification layer on \texttt{CIFAR-100}, on top of the frozen encoder. We compare against adversarial self-supervised baselines, ACL~\cite{Jiang2020PretrainingACL} and ROCL~\cite{Kim2020AdversarialCL}. As shown in Table~\ref{tab:transfering}, \modelname achieves better accuracy across all cases.
\begin{table}[t!]
\caption{Comparison in a transfer learning setup across the \texttt{CIFAR-10} and \texttt{CIFAR-100} datasets with ResNet-18. Column ``Source'' corresponds to the dataset the encoder model is trained on, while ``Target'' presents the dataset used for evaluating the frozen encoder with an additional randomly-initialized finetuned linear layer (linear evaluation).}
\label{tab:transfering}
\centering
\resizebox{\textwidth}{!}{%
\begin{tabular}{l|ccc|ccc}
\toprule
Model & Source & Target & $A_{nat}$ & Source & Target & $A_{nat}$ \\ 
\midrule\midrule
ACL~\cite{Jiang2020PretrainingACL} & \multirow{3}{*}{\texttt{CIFAR-100}} & \multirow{3}{*}{ \texttt{CIFAR-10}} & 70.08 & \multirow{3}{*}{\texttt{CIFAR-10}} & \multirow{3}{*}{ \texttt{CIFAR-100}} & 42.29 \\
RoCL~\cite{Kim2020AdversarialCL}  &  &  & 73.93 &  &  & 45.84 \\
\textbf{\modelname (Ours)}  &  &  & \textbf{75.52} &  &  & \textbf{46.01} \\ \bottomrule
\end{tabular}%
}
\end{table}

\subsection{Ablation Studies}
We study the trade-off between robustness and accuracy. In particular, we design an ablation study among our proposed \modelname and a stochastic variant that selects between the targeted attacks and the traditional adversarial contrastive learning that maximizes the loss between positive pairs. Since our method improves accuracy on clean images and on unseen $\ell_1$ attacks, we hope that this stochastic selection will be able to balance between this improvement and robustness to seen attacks. In Table~\ref{tab:ablation_prob_selection}, the first column $p$ presents the probability of selecting an adversarial generation strategy. For example, $p=0.90$ denotes applying the proposed \modelname adversarial contrastive method to $90\%$ of the batches during training, while applying RoCL to $10\%$ of the batches. Results show that smaller $p$ values result in improvements on seen $\ell_\infty$ attacks and unseen $\ell_2$ attacks, while the performance on clean images and unseen $\ell_1$ attacks improves as $p$ increases. We leave further exploration with other stochastic variants and extensive hyper-parameter tuning to future work. We perform additional ablation studies to test the effect of clustering and the variability on performance as training progresses. Results can be found in the supplementary material.

\begin{figure*}[t!]
    \centering
    \includegraphics[width=\textwidth]{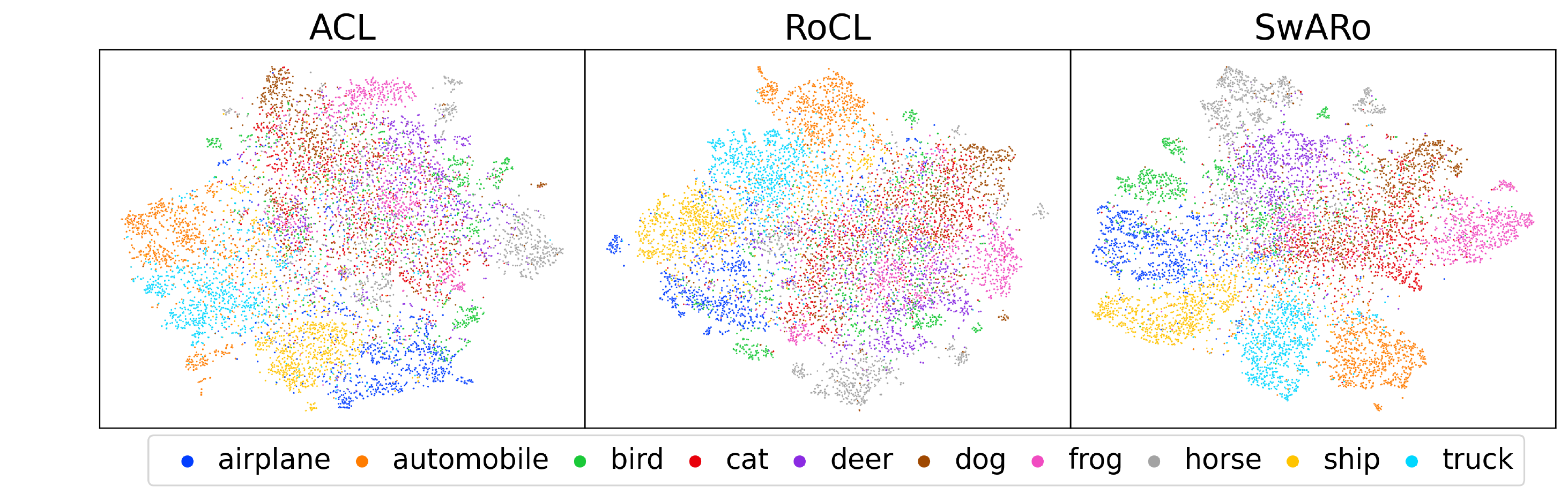}
    \vspace{-0.2cm}
    \caption{{t-SNE visualization of the three contrastive pretraining methods ACL~\cite{Jiang2020PretrainingACL}, RoCL~\cite{Kim2020AdversarialCL} and \modelname (Ours) on the \texttt{CIFAR-10} dataset. The 10 classes are represented with different colors, and the legend shows label information. \modelname
    produces well-separated clusters compared to the rest of the baselines.}}
    \label{fig:tsne}
\end{figure*}
\begin{table}[t!]
\caption{Performance of \modelname, varying the probability of selection $p$ for the proposed adversarial example generation or reverting to the traditional contrastive adversarial example generation~\cite{Kim2020AdversarialCL}. Results reported on the \texttt{CIFAR-10} dataset.}
  \label{tab:ablation_prob_selection}
   \centering
  \resizebox{.7\columnwidth}{!}{
{
  \begin{tabular}{cccccccc} 
    \toprule
    \multirow{2}{*}{\makecell{\\\\ p}} 
    &{}& \multicolumn{2}{c}{\emph{seen}} & \multicolumn{4}{c}{\emph{unseen}} \\
    \cmidrule(r){3-4} \cmidrule(r){5-8}  
    {}&{}&\multicolumn{2}{c}{$\ell_{\infty}$}&\multicolumn{2}{c}{$\ell_{2}$}&\multicolumn{2}{c}{$\ell_{1}$}\\
    \cmidrule(r){3-4} \cmidrule(r){5-6} \cmidrule(r){7-8} 
    {}&{$A_{nat}$}& $\epsilon$ 8/255 & 16/255 &0.25&0.5&7.84&12   \\
    \midrule \midrule
    0.10 & 84.11 & 39.58 & 10.68 & 73.80 & 72.51 & 82.31 & 82.25\\
    0.25 & 84.11 & 37.93 & 9.26 & 73.09 & 70.36 & 82.04 & 81.92\\
    0.50 & 85.78 & 38.37 & 10.10 & 68.78& 67.76 & 83.30 & 82.70\\
    0.75 & 87.38 & 37.33 & 9.13 & 59.69 & 59.10 & 84.06 & 83.69 \\
    0.90 & 89.13 & 35.95 & 9.68 & 42.02 & 47.86 & 84.98 & 84.36\\
    \bottomrule
  \end{tabular}
  }
  }
  \vspace{-0.3cm}
\end{table}

\subsection{Qualitative Analysis}
We present qualitative results on the learned representations. We create t-SNE visualizations~\cite{van2008visualizing} for RoCL, ACL and our method \modelname. Each semantic class on the \texttt{CIFAR-10} dataset is assigned a different color, and the legend presents the object category.
Figure~\ref{fig:tsne} shows that representations obtained from \modelname produce well-separated clusters compared to the baselines. Overall our results indicate that \modelname results in better representation learning performance and improved robustness to unseen attacks.

\section{Conclusion}\label{sec:conlusion}
In this work, we present \modelname, an adversarial contrastive learning method that learns robust self-supervised representations. Our approach is able to better guide the adversarial perturbation process. More specifically, by permuting pseudo-labels our method is able to create targeted attacks that maximize the loss for positive pairs and minimize the loss for negative pairs, leading to improvements in clean accuracy and unseen attacks. 
Through experimental analysis on black-box and white-box attacks, with two benchmark datasets and comparing against a variety of baselines, ranging from vanilla self-supervised methods to previously proposed adversarial contrastive learning approaches, we showcase that our proposed approach, \modelname, outperforms baselines that only consider positive pairs when creating adversarial examples. Additional experiments on a diverse set of white-box attacks show that \modelname is robust on realistic adversarial scenarios against a diverse set of attack types. 
Qualitative analysis shows that incorporating pseudo-label information produces well-separated clusters, leading to better learned robust representations.
In the future, we hope to evaluate our method on additional datasets with more complex model architectures. Future directions could include extensions to multi-modal data and structured tasks, as well as improving robustness on hierarchical fine-grained representation learning settings. 

\bibliographystyle{plainnat}
\bibliography{biblio}

\newpage
\appendix

\section{Appendix}
\subsection{Training Progress}
In addition, we present the learning progress as the number of epochs increases. Table~\ref{tab:ablation_epochs} reports results on \texttt{CIFAR-10}, on both clean accuracy and robust accuracy on seen and unseen attacks. 
\begin{table}[h!]
\caption{Ablation study on the number of epochs. Results reported on the \texttt{CIFAR-10} dataset.}
  \label{tab:ablation_epochs}
  \centering
{ 
  \begin{tabular}{cccccccc} 
    \toprule
    \multirow{3}{*}{\makecell{\\\\Number of Epochs}} 
    &{}& \multicolumn{2}{c}{\emph{seen}} & \multicolumn{4}{c}{\emph{unseen}} \\
    \cmidrule(r){3-4} \cmidrule(r){5-8}  
    {}&{}&\multicolumn{2}{c}{$\ell_{\infty}$}&\multicolumn{2}{c}{$\ell_{2}$}&\multicolumn{2}{c}{$\ell_{1}$}\\
    \cmidrule(r){3-4} \cmidrule(r){5-6} \cmidrule(r){7-8} 
    {}&{$A_{nat}$}& $\epsilon$ 8/255 & 16/255 &0.25&0.5&7.84&12   \\
    \midrule \midrule
    200 & 78.03 & 27.73 & 6.44 & 51.96 & 51.66 & 78.14 & 77.64\\
    400 & 81.97 & 23.90 & 4.69 & 51.87 & 50.90 & 73.46 & 72.99\\
    600 & 83.32 & 29.79 & 6.31 & 54.82 & 53.20 & 79.85 & 79.34\\
    800 & 85.63 & 32.04 & 6.51 & 59.78 & 58.90 & 81.67 & 81.25\\
    1000 & 84.43 & 33.03 & 7.52 & 57.65 & 57.20 & 82.52 & 81.99\\
    1200 & 86.55 & 33.91 & 7.92 & 61.13 & 59.92 & 83.44 & 82.98\\
    1400 & 87.05 & 36.53 & 9.13 & 59.38 & 59.35 & 	83.97 & 83.57\\
    1600 & 87.21 & 36.32 & 8.62 & 60.66 & 60.50 & 84.23 & 83.82\\
    1800 & 87.35 & 36.94 & 8.80 & 58.81 & 59.21 & 84.09 & 83.69\\
    2000 & 87.38 & 37.33 & 9.13 & 59.69 & 59.10 & 84.06 & 83.69 \\
    \bottomrule
  \end{tabular}
  }
\end{table}

\subsection{Number of Clusters}
Table~\ref{tab:ablation_num_clusters} presents results with varying the number of clusters $K$. In general, there are marginal performance differences across clean and robust accuracy, and seen and unseen attacks, showing that \modelname remains robust to the choice of this hyper-parameter.
\begin{table}[h!]
\caption{Ablation study on the number of clusters. Results reported on the \texttt{CIFAR-10} dataset. }
  \label{tab:ablation_num_clusters}
  \centering
  \resizebox{.8\textwidth}{!}{
{
  \begin{tabular}{cccccccc} 
    \toprule
    \multirow{3}{*}{\makecell{\\\\ \#$K$ Clusters}} 
    &{}& \multicolumn{2}{c}{\emph{seen}} & \multicolumn{4}{c}{\emph{unseen}} \\
    \cmidrule(r){3-4} \cmidrule(r){5-8}  
    {}&{}&\multicolumn{2}{c}{$\ell_{\infty}$}&\multicolumn{2}{c}{$\ell_{2}$}&\multicolumn{2}{c}{$\ell_{1}$}\\
    \cmidrule(r){3-4} \cmidrule(r){5-6} \cmidrule(r){7-8} 
    {}&{$A_{nat}$}& $\epsilon$ 8/255 & 16/255 &0.25&0.5&7.84&12   \\
    \midrule \midrule
    25 & 87.61 & 37.34 & 9.49 & 59.07 & 59.91 & 84.36 & 84.10\\
    50 & 87.12 & 37.86 & 9.48 & 57.98 & 60.72 & 84.60 & 84.21\\
    100 & 87.42 & 37.36 & 9.61 & 58.91 & 59.41 & 84.21 & 83.79\\
    250 & 87.79 & 37.72 & 9.17 & 59.22 & 61.21 & 85.26 & 84.84\\
    500 & 87.42 & 36.98 & 9.26 & 59.12 & 59.44 & 84.37 & 83.81\\
    1000 & 87.38 & 37.33 & 9.13 & 59.69 & 59.10 & 84.06 & 83.69 \\
    2500 & 87.38 & 37.83 & 8.96 & 58.68 & 59.90 & 84.57 & 84.08\\
    5000 & 87.34 & 37.35 & 9.67 & 59.26 & 59.41 & 84.44 & 83.95\\
    10000 & 87.48 & 38.17 & 9.01 & 56.46 & 59.91 & 84.37 & 83.70\\
    \bottomrule
  \end{tabular}
  }
  }
\end{table}

\subsection{Hyper-parameter Details}
For all models, we adopt the training setup of RoCL~\cite{Kim2020AdversarialCL}. For completeness, in this supplementary material we report important hyper-parameters, as well as newly introduced hyper-parameters. We use a $0.01$ learning rate for linear evaluation and $0.02$ for robust linear evaluation. We utilize a ResNet-18 as a base encoder for all models. In the training phase, we use a two-layered projection layer as the projection head. The embedding dimension of the projection head is 128. For the adversarial attacks in the training phase, we set $\epsilon=8/255$ and $\eta=1/255$, and $t=7$. Here $\epsilon$ is the perturbation radius, $\eta$ is the step size, and $t$ is the number of PGD iterations. Unless otherwise specified, we use $K=1000$ for CIFAR-10 and $K=5000$ for CIFAR-100, where $K$ is the number of clusters. We set $p=0.75$, which is the probability of applying \modelname to a batch. We train all our models for $2000$ epochs. We use a batch size of $256$ for running all our training. For the rest, we follow a similar optimization setup and data augmentation as RoCL~\cite{Kim2020AdversarialCL}. Training time for \modelname is approximately 5 hours and 20 minutes per 100 epochs with two NVIDIA V100 GPUs with 16GB memory.

\end{document}